\documentclass[sigconf,10pt,nonacm]{acmart}
\AtBeginDocument{%
  }

\setcopyright{acmlicensed}
\copyrightyear{2018}
\acmYear{2018}
\acmDOI{XXXXXXX.XXXXXXX}
\acmConference[Conference acronym 'XX]{Make sure to enter the correct
  conference title from your rights confirmation email}{June 03--05,
  2018}{Woodstock, NY}
\acmISBN{978-1-4503-XXXX-X/2018/06}

\usepackage{float}
\floatstyle{boxed}            
\newfloat{promptbox}{tbp}{lop} 
\floatname{promptbox}{Prompt}

\newfloat{examplebox}{tbp}{lop} 
\floatname{examplebox}{Example}

\begin{document}

\title{ActionNex: A Virtual Outage Manager for Cloud Computing}

\settopmatter{authorsperrow=4}
\author{Zhenfeng Lin}
\authornote{Microsoft PRIMO.}

\author{Ryan Zhang\footnotemark[1]}

\author{Chetan Bansal}
\authornote{Microsoft Research}

\author{Salman Zafar\footnotemark[1]}

\affiliation{  \institution{\footnotemark[1]Microsoft PRIMO}
  \city{Redmond}
  \country{USA}}
\author{Haoji Hu}
\authornote{Microsoft Azure Core}

\author{Junhao Li \footnotemark[1]}

\author{Hatay Tuna \footnotemark[4]}

\affiliation{  \institution{Microsoft Research}
  \city{Redmond}
  \country{USA}}

\author{Ming Hao}
\authornote{ Microsoft Azure CTO Office}

\author{Ze Li\footnotemark[3]}

\author{Murali Chintalapati\footnotemark[3]}

\affiliation{  \institution{Microsoft Azure Core}
  \city{Redmond}
  \country{USA}}

\author{Xuchao Zhang\footnotemark[2]}

\author{Oleg Kulygin \footnotemark[1]}

\author{Sheila Jiang \footnotemark[1]}

\author{Angie Anderson \footnotemark[1]}
\affiliation{  \institution{Azure CTO Office} 
  \city{Redmond}
  \country{USA}}

\renewcommand{\shortauthors}{Lin et al.}

\begin{abstract}
Outage management in large-scale cloud operations remains heavily manual, requiring rapid triage, cross-team coordination, and experience-driven decisions under partial observability. We present \textbf{ActionNex}, a production-grade agentic system that supports end-to-end outage assistance, including real-time updates, knowledge distillation, and role- and stage-conditioned next-best action recommendations. ActionNex ingests multimodal operational signals (e.g., outage content, telemetry, and human communications) and compresses them into critical events that represent meaningful state transitions. It couples this perception layer with a hierarchical memory subsystem: long-term Key--Condition--Action (KCA) knowledge distilled from playbooks and historical executions, episodic memory of prior outages, and working memory of the live context. A reasoning agent aligns current critical events to preconditions, retrieves relevant memories, and generates actionable recommendations; executed human actions serve as an implicit feedback signal to enable continual self-evolution in a human--agent hybrid system. We evaluate ActionNex on eight real Azure outages (8M tokens, $\sim$4{,}000 critical events) using two complementary ground-truth action sets, achieving 71.4\% precision and 52.8--54.8\% recall. The system has been piloted in production and has received positive early feedback.
\end{abstract}




\maketitle

\section{Introduction}

Cloud computing has became dominant computing paradigm and continues to expand in both scale and complexity. Outage Management is a core function in large-scale cloud operations: it coordinates detection, assessment, investigation, mitigation, and resolution to minimize service disruptions and uphold strict service-level objectives. Yet, in sharp contrast to the growing automation of software creation, outage-management operations remain heavily manual—relying on human-driven triage, cross-team coordination, and experience-based decision making under time pressure. 

In the recent years, the rapid gains in LLM reasoning have enabled agentic systems to plan over long horizons, use external tools, operate under partial observability, and improve through feedback ~\cite{wei2026agenticreasoning},~\cite{zhang2025agentorchestra}, ~\cite{schick2023toolformer}, ~\cite{jin2025searchr1}, ~\cite{zeng2023agenttuning}. As a result, automating outage-management tasks is becoming increasingly reachable. Multiple systems have been proposed in this domain ~\cite{yu2025triangle}, ~\cite{fu2025oncallx}, ~\cite{mao2025agentictroubleshooting}, ~\cite{hamadanian2023holistic}, ~\cite{wang2025llmfailurelocalization}, ~\cite{lou2013softwareanalytics}, ~\cite{wang2024rcagent}.  Most of them, however, focus on one or a few aspects of outage-management activities—such as troubleshooting execution, triage, or causal reasoning.

In practice, outage management spans a much wider and more variable operational space. During large-scale outages, hundreds of responders across distinct roles may participate; the response can last days; and dozens of work-streams often run in parallel while exchanging information intensively and continuously. Modern outage-management environments therefore involve a broad range of cognitive and coordination-heavy tasks, including custom-impact analysis, monitoring and diagnosis, triage, internal and external communications, cross-service recovery orchestration, resolution validation, and so on. As a result, specialized systems typically provide only partial assistance in the real-world operations. 
 
To design a general system that can tackle these tasks end-to-end, we turn to reinforcement learning paradigm and present \textbf{ActionNex}, a continual, self-evolving outage management solution which uses action as general abstraction for knowledge, decision-making and tasks. 

The core idea is to learn what actions to take and when to take the action from human experiences while operating in a human-agent hybrid system so that it can recommend actions in a timely fashion (and execute them in the future).

In real time, ActionNex autonomously ingests multi-source signals—including structured data from outage systems, bridge meeting transcripts, groups chats, and other operational artifacts—and distills into actionable knowledge through hierarchical abstraction in its memory. Conditioned on outage context, stage, and role, it proposes next-best actions.  Meanwhile, the actions ultimately taken by humans act as an implicit reward signal, enabling to continuously learn, adapt, and improve over time. 
\begin{figure}[t]
  \centering
  \includegraphics[width=0.95\columnwidth]{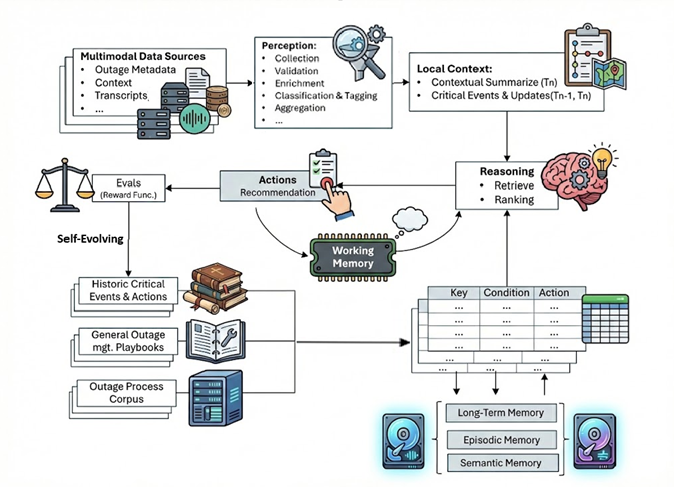}
  \caption{ActionNex Framework}
  \label{fig:framework}
\end{figure}

The architecture of ActionNex is shown at Fig.~\ref{fig:framework}. It consists of following key components:
\begin{itemize}
\item {\texttt{Mutlimodal Data Perception Layer}}: Ingests multimodal signals from diverse sources and encodes them into critical events that compactly represent and summarize outage symptoms and state transitions. 
\item {\texttt{Knowledge Memory Layer}}: Stores the extracted knowledge as \textit{(key, condition, action}) entries derived from historical human operations and playbooks. It continuously accumulates new knowledge extracted from outage management processes and self-evolves.
\item {\texttt{Reasoning Agent Layer}}: Applies policy and/or learned reasoning to align current critical events with the preconditions and adaptively generate most likely next-best actions recommendation.
\end{itemize}

We validate ActionNex with real outage data from Azure, demonstrating the effectiveness of our continual self-evolution design. Specifically, we extract ground-truth actions from raw operational traces in offline batches using latest LLM, and manually label for quality assurance. We then replay traces to trigger action recommendations, which are evaluated against the ground truth over subsequent time windows. The system has been deployed in production piloting and has received positive early feedbacks.

In summary, our contributions include:
\begin{enumerate}
    \item We design and deploy a general production-grade agentic outage management system that supports Q\&A, real-time updates, knowledge distillation, and Key-Condition-Action (\textbf{KCA}) based real-time  recommendations.
    \item  We propose a novel self-evolving  hierarchical memory subsystem.
    \item  We evaluate the system with real Azure production outage data and demonstrate its effectiveness.
\end{enumerate}

\section{Mutlimodal Data Perception Layer}

The Multimodal Data Perception Layer constitutes the front-end of the ActionNex. It transforms diverse, high-volume signals into a human-readable compact representation -- critical events -- for downstream reasoning and action recommendation instead of exposing noisy raw input directly to the reasoning agent. These events summarize the system state and the outage investigation evolution. This transformation not only improves the detection of operationally salient state changes, but also allow the whole  process to be interpretable and easy to maintain. 

\subsection{Input Modalities}

The system ingests multiple complementary data modalities that jointly characterize the evolving state of  outage:
\begin{itemize}
    \item Outage-management operational metadata (e.g., affected services/regions, timestamps, severity indicators, structured status updates, communications). 
    \item Human communication artifacts (bridge meeting transcripts, chats, operator notes revealing hypotheses, decisions, ownership, and implicit intent).
    \item Telemetry-derived contextual signals (alerts, anomaly annotations, partner events reflecting system changes and mitigation effects). 
\end{itemize}
Each modality is a partial and noisy on its own, so effective action recommendation requires joint interpretation across modalities rather than isolated signal processing.

\subsection{Normalization and Perception}

Incoming signals are first processed through a perception pipeline that emphasizes semantic alignment over raw fidelity. Its output is then used to generate critical events.

First, inputs are normalized into a common representation, validated for temporal consistency and source authenticity, and deduplicated to remove redundant observations. Next, the system enriches signals by extracting entities (e.g., services, regions, components, teams), resolving aliases, and augmenting observations with dependency and topology context. Signals are tagged and classified by outage phase, event type, severity, and relevance to support focused downstream reasoning. Finally, cross-modalities signals are aggregated into coherent contextual states representing the system’s current understanding of the outage. 

This pipeline deliberately avoids early decision-making. Its reduces heterogeneity while preserving the semantic content needed to reason about operational impact.

\subsection{Critical Event Abstraction}

\noindent
Reasoning directly over raw multimodal outage data is neither scalable nor aligned with operational practice. Signals are high-volume, noisy, and often redundant, and decisions are rarely triggered by individual alerts; they respond to meaningful changes in system state.  We therefore use critical events as principled abstraction of state of change. By collapsing observations into explicit state transitions, the perception layer reduces noise, captures temporal structure, and aligns perception with human outage-manager mental models.

This layer outputs a sequence of critical events—observations or transitions that materially alter outage understanding or control (e.g., scope expansion, mitigation start/complete, dependency recovery, probable root cause identification). The system computes deltas between consecutive time windows and promotes significant deltas to critical events, storing them in short-term memory as the interface to downstream reasoning.


Although critical events are a lossy abstraction, they are sufficient for action recommendation because they retain decision-critical state changes (e.g., mitigation progress, dependency recovery, blast-radius expansion), are derived from aggregated cross-modal enriched context (topology, life-cycle phase, corroboration), and provide the predicates required by ActionNex’s KCA reasoning without exposing raw noise. 
\begin{table*}[t]
\centering
\small
\setlength{\tabcolsep}{6pt}
\renewcommand{\arraystretch}{1.25}
\begin{tabular}{p{3.2cm} p{7.6cm} p{5.8cm}}
\hline
\textbf{Memory} & \textbf{Source \& Content} & \textbf{Role in Reasoning} \\
\hline

\textbf{Long-Term Memory} \newline (Knowledge Base)
&
Aggregated reference knowledge, including outage management manuals, SOPs, service \textbf{playbooks}, guidelines, and domain-specific expertise, indexed as \emph{Key--Condition--Action} (KCA) triples.
&
Provides \textbf{canonical solutions} and best-practice actions for known scenarios when matching conditions are satisfied.
\\

\textbf{Episodic Memory} \newline (Case Library)
&
Logs of \textbf{historical outages}, including critical events, decisions taken, and outcomes (success or failure). Each case records the outage context (e.g., service, severity), the identified precondition, and the actions used for resolution.
&
Provides \textbf{analogous examples} from prior outages. Successful actions suggest viable strategies, while failed actions indicate approaches to avoid.\\

\textbf{Working Memory} \newline (Current Context)
&
\textbf{Live Outage state}, including recently attempted actions, intermediate results, conversation history, and the current outage stage or status. This memory is continuously updated as the outage evolves.&
Provides \textbf{situational awareness}, ensuring recommendations account for what has already occurred, thereby preventing redundant or contradictory actions and maintaining a coherent action sequence.
\\

\hline
\end{tabular}
\caption{Memory types used by the agent and their roles in outage reasoning and action recommendation.}
\label{tab:memory_types}
\end{table*}
\section{Knowledge Memory Layer} \label{sec:memory}

The goal of memory design is to enable effectively information learning, indexing, and retrieval for relevant action recommendations. There are a large body of research demonstrating the essential role memory plays for self-evolving \cite{cao2025rememberme}, \cite{anwar2024remembr}, \cite{wei2025evomemory}, \cite{ouyang2025reasoningbank}, \cite{anonymous2026remem}, \cite{xia2026memora}. Our knowledge memory layer is a hieratically structure consisting of multiple memory blocks shown in the table~\ref{tab:memory_types} which capture complementary forms of context (e.g., recent outage state, historical trajectories, and reusable operational knowledge). 

These components together accumulate knowledge and guide recommendations. Due to space constraints, we will give a brief introduction on working and episodic memory first and focus on our core component—long-term memory afterwards. Working memory tracks the live outage state (recent actions, results, stage) to prevent redundant or conflicting steps. Episodic memory stores past outage cases (critical events, decisions, outcomes) to reuse proven strategies and avoid failures. Over time, both support continuous improvement. 

The main functionality of long-term memory is to connect perceived outage dynamics to operational actions by organizing knowledge into structured key, condition, action (KCA) units for retrieval, condition evaluation, and policy grounding. Because KCA elements are stored as text, they are interpretable and easy for domain experts to audit and correct; domain experts can also directly author KCA entries, enabling learning from both real-world outage data and human knowledge.  This design lets the reasoning agent match the current outage state against accumulated experience and recommend actions aligned with operational practice. The KCA definitions are shown below:

\textbf{Condition:}
Summarizes generalized symptoms distilled from critical events (e.g., root cause confirmed, mitigation in progress, dependency recovered, or escalation triggered). Each condition acts as a stable semantic identifier that consolidates multiple observations of the same operational phenomenon.

\textbf{Key:}
Provides the contextual scopes for interpreting a condition, typically tied to life-cycle stage, service domain,  or responsibility boundary (e.g., investigate/mitigate/recover). Keys and conditions are embedded and indexed; either may surface KCA candidates first, with the other used for filtering or ranking, enabling robust retrieval under partial or uncertain context.

\textbf{Action:}
Specifies the operational response once a KCA unit is retrieved and its context is deemed relevant. Actions are not primary retrieval targets during recommendation reasoning progress; they are normalized from historical executions and established procedures, represented as parameterized semantic templates that adapt to specific services/regions/roles while preserving real operator structure for precision, auditability, and trust. 

In operation, each KCA unit functions as a contextual implication: under a given key, if its condition is supported by the evolving critical events, the linked action becomes a candidate recommendation. The reasoning agent embeds the current outage state and retrieves relevant units via similarity of keys or conditions, then incorporates the remaining context to refine relevance, rank alternatives, and suppress unsafe or mistimed responses. Because selection is retrieval-driven rather than dependent on exact rule triggering, the framework naturally accommodates partial evidence, ambiguity, and incremental updates as new information arrives. Executed outcomes are continuously written back into memory, enabling the knowledge base to evolve while remaining anchored to canonical and auditable forms. 

Overall, KCA  provides a principled bridge between perception and decision-making, balancing abstraction with flexibility while preserving alignment with how human outage managers interpret situations and authorize actions.

All memory components support continual updates implementing implicit self-evolution. In online fashion, the system selectively  promotes working memory traces into  episodic memory, and periodically summarizes new knowledge and merge into long-term memory. In offline fashion, domain experts can optionally review memory records and provide update recommendations, which are incorporated to evolve the whole memory system.



\section{Reasoning Agent Layer}

The Reasoning Agent Layer is the decision-making core of ActionNex. It bridges perceived critical events and the memory layer (Section~\ref{sec:memory}) to produce action recommendations. This layer interprets incoming critical events to infer the current outage situation as a precondition. It then combines this inferred context with retrieved long-term, episodic, and working memories to recommend the next action under its policy. 

\begin{center}
\noindent
\fbox{%
  \begin{minipage}{0.8\columnwidth}
    \footnotesize
    \textbf{Example}\\[-2pt]
    \rule{\columnwidth}{0.3pt}
    \textit{
    \textbf{04:19Z:}
    BridgeVoice: “thermal protection… triggered shutdowns” of “storage and direct 
    drive,” while “cooling… restored” and “temperatures… stabilizing” under 
    “SEV-1.” The system matched the Resolve precondition for “Drive Core Platform 
    Recovery Sequence.”\\
    \textbf{04:22Z:}
    recommended “controlled power-up and recovery sequence” for the “shutdown 
    direct drive cluster.”\\
    \textbf{04:37Z:}
    humans executed “Controlled, staged recovery initiated” and “completed… 
    recovery sequencing plans.”
    }
  \end{minipage}
}
\end{center}

\begin{description}
    \item[\textbf{Step 1}]
    \textbf{Align Events to Preconditions.}
    The agent converts raw outage events at time $t$ into a structured precondition describing the current situation, along with meta-context, such as outage stage and affected service. 

    \item[\textbf{Step 2}]
    \textbf{Retrieve Relevant Memories.}
    The agent retrieves the top-$k$ entries from long-term memory and the top-$m$ similar cases from episodic memory, incorporating working-memory context when available.

    \item[\textbf{Step 3}]
    \textbf{Adaptive Action Recommendation with Feedback.}
    Given the precondition and retrieved memories, the agent formulates a set of candidate actions and uses its policy—implemented via an LLM prompt—to recommend the next best action. If the LLM determines that the available context is insufficient (e.g., missing key signals or conflicting evidence), the agent autonomously re-enters Step 2, refining its memory queries or expanding the retrieval scope to gather additional relevant information. This refine process is similar to the refine process in React~\cite{yao2022react} to fine-tune the prompt and regenerate the recommendation results. Prompt~\ref{box:system_prompt} illustrates the action recommendation step.
 \end{description}

\setlength{\fboxsep}{6pt}
\setlength{\fboxrule}{0.6pt}

\begin{promptbox}[tb]  
\caption{Simplified prompt for one action recommendation attempt}\label{box:system_prompt}
\raggedright
\textbf{System.} You are a reasoning agent that recommends outage mitigation actions
based on structured context and operational knowledge.

\vspace{0.3em}
\textbf{Situation (Precondition).}
Power and cooling have been restored after automatic thermal-protection shutdowns
impacting storage and direct-drive components.
All dependencies are verified and mitigation is complete
(\emph{Stage}: Mitigate; \emph{Outage}: ``High Thermal Triggered Shutdowns'').

\vspace{0.3em}
\textbf{Recent Observations (Working Memory).}
At 04{:}19Z, bridge updates reported thermal-triggered shutdowns.
Subsequent communications confirmed cooling restoration and temperature
stabilization.
Outage severity is \texttt{SEV-1}.

\vspace{0.3em}
\textbf{Relevant Knowledge (Long-Term).}
\begin{itemize}
    \item \emph{Runbook}: After thermal shutdowns, verify environmental stability
    before initiating the \textbf{Drive Core Platform Recovery Sequence}.
    \item \emph{KCA}: Ensure mitigation and dependency health prior to controlled
    service restoration.
\end{itemize}

\vspace{0.3em}
\textbf{Similar Past Outages (Episodic Memory).}
\begin{itemize}
    \item \emph{Outage \#312} (Aug 2024): Thermal event caused storage shutdowns.
    \textbf{Action}: Cooling restored, followed by staged platform recovery.
    \item \emph{Outage \#198} (Jan 2023): Protective shutdowns due to high temperature.
    \textbf{Action}: Environmental stabilization before recovery.
\end{itemize}

\vspace{0.3em}
\textbf{Task.}
Given the current precondition, observations, and knowledge,
\textbf{recommend the next mitigation or recovery action}.
\end{promptbox}

The next section will evaluate how this reasoning approach performs in practice and how it interacts with the rest of the system (including any learning or feedback mechanisms for refining the policy).

\section{Evaluation Results}
\label{sec:evaluation}

We evaluate recommendation quality, system efficiency, and contribution of key components. Our evaluation is designed to reflect both the noise of real outage response and alignment with high-confidence operational guidance.

\textbf{Dataset Overview: }
We curated a dataset consisting of eight real-world outages, comprising approximate 8M tokens of contextual data, including outage, operational logs, meeting transcripts, and group chats. From raw contexts, we extracted 4{,}374 critical events, serving as the intermediate state representation for downstream reasoning and led to 361 recommended actions. In parallel, 447 actions were independently identified and confirmed by human investigators and treated as ground-truth labels, without reference to the model predictions. Five outages were used for memory construction and development, while the remaining three were held out exclusively for evaluation.

\textbf{LLM Models usage:}
Throughout this work, we employ \textbf{GPT‑5.2} for information extraction, memory construction, and multi‑step reasoning. We use \textbf{text-embedding‑3‑large} for embedding generation and memory indexing. For final prediction–-ground‑truth evaluation, we rely on \textbf{GPT‑5.2 Thinking}, which provides higher‑fidelity judgment for outcome assessment.

\subsection{Ground Truth Construction}
\label{sec:gt}
 To enable reliable evaluation, we curate two complementary ground truth sets. We further invited domain experts to manually evaluate them and general feedback is that they cover almost all the common actions for outage management activities ignoring outage specific details.

\textbf{G1 (In-the-wild actions):}  
G1 consists of actions extracted directly from outage context. This set is closest to real deployment conditions, but noisy and occasionally ambiguous, as actions may be discussed without execution or logged without explicit attribution.

\textbf{G2 (Traceable guidance actions):}  
G2 is derived by filtering G1 to retain only actions explicitly supported by authoritative playbooks or handbooks. Relative to G1, G2 is more concise and easier to verify, focusing on higher-confidence, repeatable operational patterns.

Evaluating against both G1 and G2 allows us to disentangle two questions: (I) Robustness under realistic operational noise (G1), and (II) Correctness with respect to canonical, high-confidence guidance (G2).

\subsection{Data Quality and Semantic Coverage}
\label{sec:coverage}

\begin{figure}[t]
    \centering
    \includegraphics[width=\columnwidth]{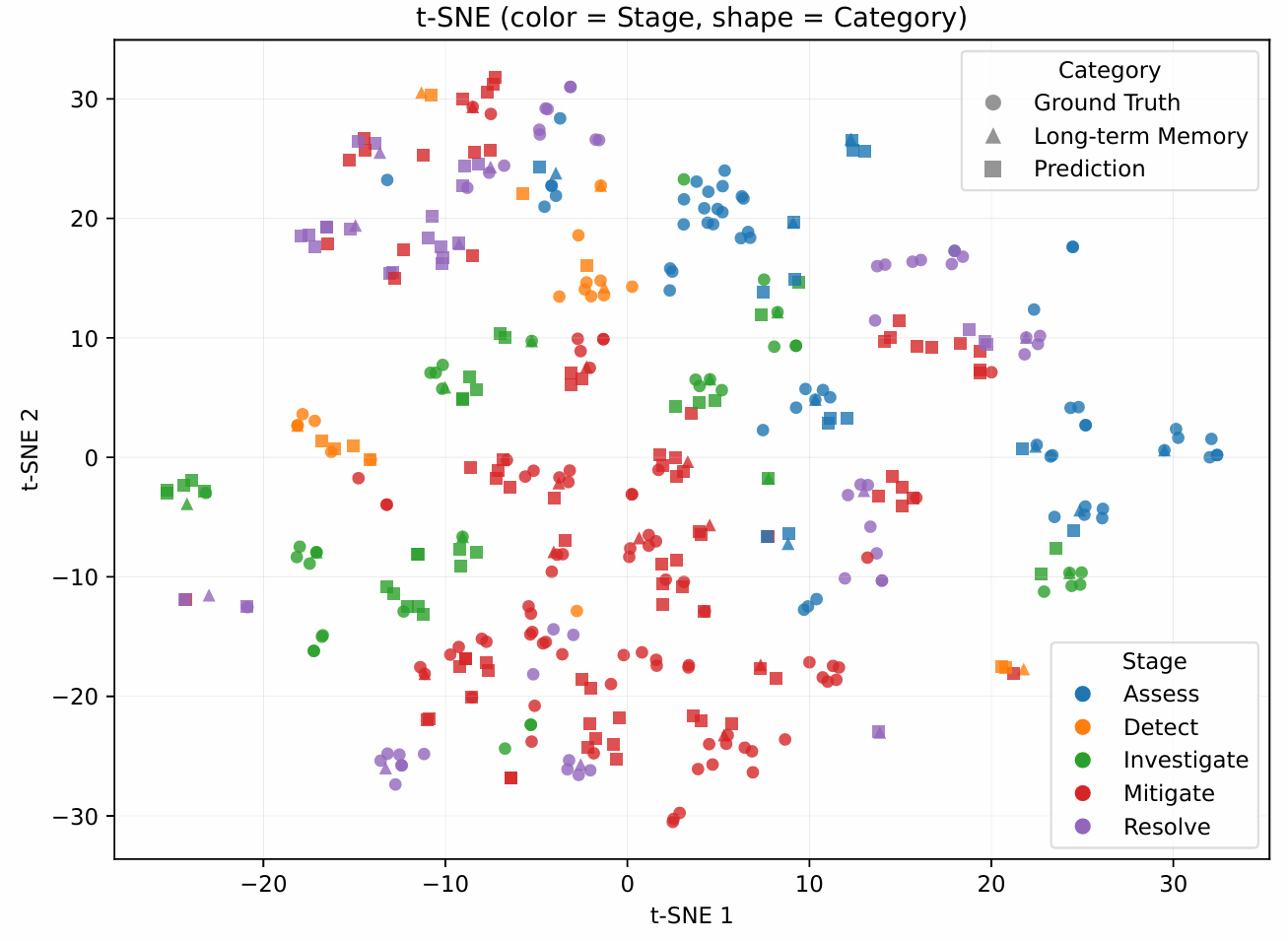}
    \caption{t-SNE view of actions}
    \label{fig:semantic_coverage}
\end{figure}

To validate the semantic quality of both ground-truth sets and ensure meaningful metrics, we perform a coverage analysis using embedding-space visualization. For each outage, we embed predicted and ground truth actions with outage life-cycle stage tags, and visualize them using t-SNE projections. Figure~\ref{fig:semantic_coverage} shows predicted actions (square), G1 actions (circle) and playbook-extracted long-term memory actions (triangle).

Both sets, predicted actions broadly and evenly cover the same semantic regions occupied by other actions. This indicates that (I) the curated ground truth sets occupy coherent semantic regions rather than isolated artifacts, and (II) ActionNex generates a diverse yet relevant set of actions aligned with human operational actions. This supports that the reported precision/recall reflects real semantic alignment rather than extraction bias.

\subsection{Performance}
\label{sec:perf}

\begin{table}[t]
\centering
\small
\setlength{\tabcolsep}{4pt}
\renewcommand{\arraystretch}{0.95}
\begin{tabular}{l c c}
\hline
\textbf{Ground Truth} & \textbf{Precision (\%)} & \textbf{Recall (\%)} \\
\hline
G1 & \textbf{71.4} & \textbf{52.8} \\
G2 & 71.4          & 54.8          \\
\hline
\end{tabular}
\caption{Accuracy for G1 and G2.}
\label{tab:vom_accuracy}
\end{table}

Table~\ref{tab:vom_accuracy} summarizes precision and recall evaluated against two ground‑truth sets, \textbf{G1} and \textbf{G2}. Overall, results suggest ActionNex can learn operational actions from human experiences and playbooks and surface them at appropriate time given evolving outage context. Precision happen to be identical for both cases because matching ground truth are preserved when generating \textbf{G2} from \textbf{G1}. While this precision might appear modest, manual inspection reveals that majority of false positives actually are operationally reasonable recommendations that were simply not captured or promoted into ground truth. Relatively low recall is mainly because limited early stage outage context delays confident triggering, or related conditions are not identified by the system yet. ActionNex’s ability to incorporate execution outcomes and feedback over time provides a natural path to improving both coverage and timing as additional outage are observed.    

\begin{figure}[t]
    \centering
    \includegraphics[width=\columnwidth]{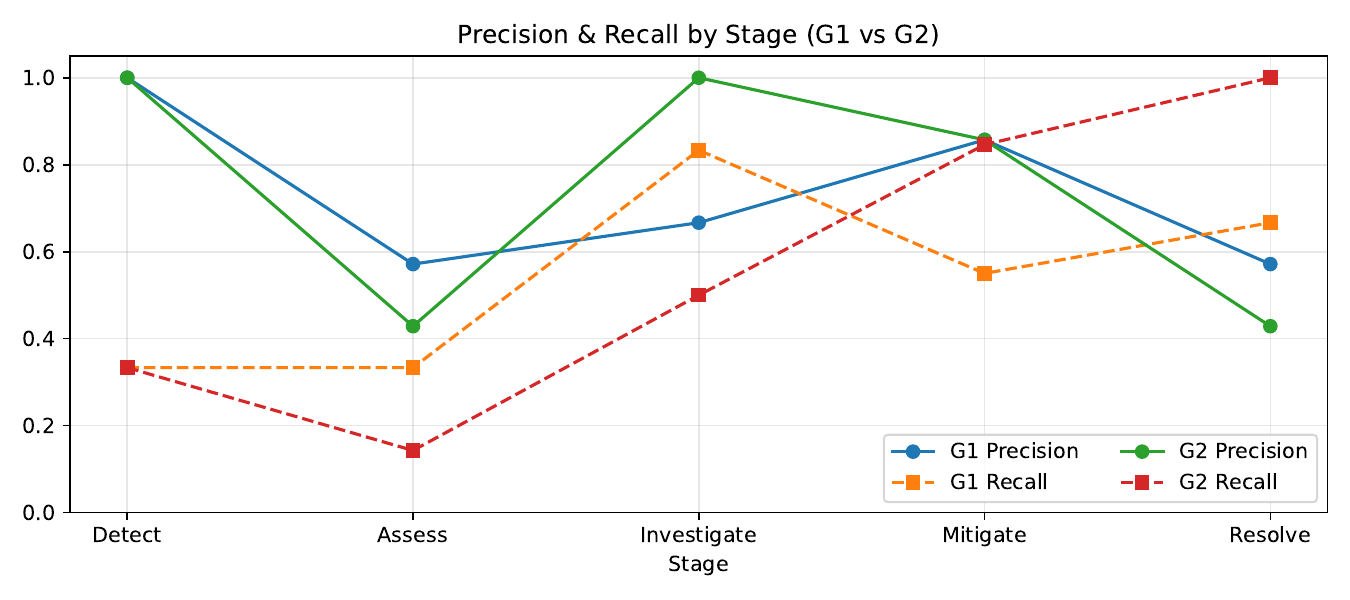}
    \caption{Accuracy over Stages for G1 and G2}
    \label{fig:precision_recall_by_stage}
\end{figure}

To understand recommendation quality evolves over time, we analyze performance across five outage stages. Figure~\ref{fig:precision_recall_by_stage} shows \textit{Stage-wise precision and recall}. Recall increases while precision decreases as outage progresses. Early stages contain spares and uncertain signals, so recommendations are constrained by incomplete context and hypotheses. As additional critical events are confirmed and outcomes become more observable, preconditions can be inferred more reliably, expanding coverage and improving recall. From product perspective, this pattern propose a stage-adaptive strategy—for instance, \textit{emphasizing episodic-memory analogies in early stages to improve recall, and shifting toward stricter condition gating later to preserve precision.}

\section{Conclusions}
\label{sec:conclusions}

We presented a production-grade agentic system for outage management that combines multimodal context, critical-events, and action recommendations with a hierarchical memory layer centered on long-term Key--Condition--Action and continual learning from human actions. On real production outages, ActionNex demonstrates strong recommendation quality and practical utility in a human-agent workflow. Future work will improve early-stage coverage, strengthen safety for online learning, and measure impact on operational outcomes such as time-to-mitigation at scale.


\bibliographystyle{ACM-Reference-Format}
\bibliography{references}

@article{xia2026memora,
  author  = {Xia, Menglin et al.},
  title   = {{Memora}: A Harmonic Memory Representation Balancing Abstraction and Specificity},
  journal = {arXiv:2602.03315},
  year    = {2026}
}

@inproceedings{yu2025triangle,
  author    = {Yu, Zhaoyang et al.},
  title     = {Triangle: Empowering Incident Triage with Multi-Agent Systems},
  booktitle = {ASE},
  year      = {2025}
}

@inproceedings{fu2025oncallx,
  author    = {Fu, Ruowei et al.},
  title     = {OncallX: {LLM}-Powered Multi-Agent Collaboration for On-Call Automation},
  booktitle = {ASE},
  year      = {2025}
}

@article{mao2025agentictroubleshooting,
  author  = {Mao, Jiacheng et al.},
  title   = {Agentic Troubleshooting Guide Automation for Incident Management},
  journal = {arXiv:2510.10074},
  year    = {2025}
}

@inproceedings{hamadanian2023holistic,
  author    = {Hamadanian, Pouya et al.},
  title     = {A Holistic View of {AI}-Driven Network Incident Management},
  booktitle = {HotNets},
  year      = {2023}
}

@inproceedings{wang2025llmfailurelocalization,
  author    = {Wang, Chenxu et al.},
  title     = {Towards {LLM}-Based Failure Localization in Production Networks},
  booktitle = {SIGCOMM},
  year      = {2025}
}

@inproceedings{lou2013softwareanalytics,
  author    = {Lou, Jian-Guang et al.},
  title     = {Software Analytics for Incident Management of Online Services},
  booktitle = {ASE},
  year      = {2013}
}

@inproceedings{wang2024rcagent,
  author    = {Wang, Zefan et al.},
  title     = {{RCAgent}: Cloud Root Cause Analysis with Autonomous Agents},
  booktitle = {CIKM},
  year      = {2024}
}

@article{zeng2023agenttuning,
  author  = {Zeng, Aohan et al.},
  title   = {AgentTuning: Enabling Generalized Agent Abilities for {LLM}s},
  journal = {arXiv:2310.12823},
  year    = {2023}
}

@article{jin2025searchr1,
  author  = {Jin, Bowen et al.},
  title   = {Search-R1: Training {LLM}s to Reason with Search via Reinforcement Learning},
  journal = {arXiv:2503.09516},
  year    = {2025}
}

@article{schick2023toolformer,
  author  = {Schick, Timo et al.},
  title   = {Toolformer: Language Models Can Teach Themselves to Use Tools},
  journal = {arXiv:2302.04761},
  year    = {2023}
}

@article{zhang2025agentorchestra,
  author  = {Zhang, Wentao et al.},
  title   = {AgentOrchestra: Orchestrating Multi-Agent Intelligence},
  journal = {arXiv:2506.12508},
  year    = {2025}
}

@article{wei2026agenticreasoning,
  author  = {Wei, Tianxin et al.},
  title   = {Agentic Reasoning for Large Language Models},
  journal = {arXiv:2601.12538},
  year    = {2026}
}

@article{cao2025rememberme,
  author  = {Cao, Zouying et al.},
  title   = {Remember Me, Refine Me: Dynamic Procedural Memory for Agent Evolution},
  journal = {arXiv:2512.10696},
  year    = {2025}
}

@article{anwar2024remembr,
  author  = {Anwar, Abrar et al.},
  title   = {ReMEmbR: Long-Horizon Spatio-Temporal Memory for Robot Navigation},
  journal = {arXiv:2409.13682},
  year    = {2024}
}

@article{wei2025evomemory,
  author  = {Wei, Tianxin et al.},
  title   = {Evo-Memory: Benchmarking Test-Time Learning with Self-Evolving Memory},
  journal = {arXiv:2511.20857},
  year    = {2025}
}

@article{ouyang2025reasoningbank,
  author  = {Ouyang, Siru et al.},
  title   = {ReasoningBank: Scaling Agent Self-Evolution with Reasoning Memory},
  journal = {arXiv:2509.25140},
  year    = {2025}
}

@inproceedings{anonymous2026remem,
  author    = {{Anonymous}},
  title     = {{REMem}: Reasoning with Episodic Memory in Language Agents},
  booktitle = {ICLR},
  year      = {2026}
}

@inproceedings{yao2022react,
  title={React: Synergizing reasoning and acting in language models},
  author={Yao, Shunyu and Zhao, Jeffrey and Yu, Dian and Du, Nan and Shafran, Izhak and Narasimhan, Karthik R and Cao, Yuan},
  booktitle={The eleventh international conference on learning representations},
  year={2022}
}










\end{document}